\documentclass[conf]{new-aiaa}
\usepackage[utf8]{inputenc}
\usepackage{graphicx,color}
\usepackage{amsmath}
\usepackage[version=4]{mhchem}
\usepackage{siunitx}
\usepackage{longtable,tabularx}
\usepackage{subcaption}
\usepackage{hyperref} 
\usepackage{arydshln}

\usepackage{optidef} 
\usepackage[ruled,vlined]{algorithm2e}  
\usepackage{mathrsfs}   
\usepackage{changepage}   
\usepackage{float} 
\usepackage{dsfont}
\usepackage{nccmath}
\usepackage{tabularx}
\usepackage{booktabs}
\usepackage{longtable}
\usepackage{adjustbox} 
\usepackage{amsmath}               

\newcommand{\figref}[1]{Fig.~\ref{#1}}

\usepackage{bm}

\usepackage{optidef} 
\usepackage[ruled,vlined]{algorithm2e}  
\usepackage{mathrsfs}   
\usepackage{changepage}   
\usepackage{float} 
\usepackage{dsfont}

\usepackage{amsthm}
\usepackage{comment}

\title{System-of-systems Modeling and Optimization: 
An Integrated Framework for Intermodal Mobility}

\author[]{\underline{Paul Saves$^{1,2}$},  Jasper H Bussemaker$^3$, Rémi Lafage$^{1,2}$,  Thierry Lefebvre$^{1,2}$, Nathalie Bartoli$^{1,2}$, Youssef Diouane$^{4}$, Joseph Morlier$^{2,5}$}

\affil[]{$^1$ DTIS, ONERA, Université de Toulouse, Toulouse, France
\\
$^2$ Fédération ENAC ISAE-SUPAERO ONERA, Université de Toulouse, Toulouse, France
\\ 
$^3$ Institute of System Architectures in Aeronautics, German Aerospace
Center (DLR), Hein-Saß-Weg 22, Hamburg, 21129, Germany.
\\ 
$^4$ GERAD and Department of Mathematics and Industrial
Engineering, Polytechnique Montréal, Montréal, QC, Canada
\\ 
$^5$ ICA, Université de Toulouse, ISAE–SUPAERO, INSA,
CNRS, MINES ALBI, UPS, Toulouse, France 
\\
\textbf{Contact:} Paul.Saves@onera.fr,Jasper.Bussemaker@dlr.de}

\usepackage{lastpage}
\usepackage{fancyhdr}
\begin{document}

\newtheorem{assumption}{Assumption}
\newtheorem{theorem}{Theorem}
\newtheorem{corollary}{Corollary}
\newtheorem{lemma}{Lemma}

\maketitle

\begin{abstract}

For developing innovative systems architectures, modeling and optimization techniques have been central to frame the architecting process and define the optimization and modeling problems~\cite{bussemaker2024system}.
In this context, for system-of-systems the use of efficient dedicated approaches (often physics-based simulations) is highly recommended to reduce the computational complexity of the targeted applications.
However, exploring novel architectures using such dedicated approaches might pose challenges for optimization algorithms, including increased evaluation costs and potential failures. 
To address these challenges, surrogate-based optimization algorithms, such as Bayesian optimization utilizing Gaussian process models have emerged. The proposed solution relies mainly on leveraging the capabilities of ONERA and DLR software as well as other open-source solutions. In fact, for this purpose we leverage the use of the ONERA software  SEGOMOE~\cite{bartoli:hal-02149236} (Super-Efficient Global Optimization with Mixture of Experts) based on surrogate models from SMT~\cite{saves2023smt} (Surrogate Modeling Toolbox) to handle hierarchical variables. We also used the DLR software SBArchOpt~\cite{bussemaker2023sbarchopt} to interface optimizers with architecture models like OpenTurbofanArchitecting~\cite{bussemaker2021system}. 
The hierarchical variables support in SMT enables an effective representation of design decisions, enhancing optimization outcome. In particular, this paper demonstrates SEGOMOE capabilities to help solving realistic architecture aircraft engine test problems that are defined within the DLR OpenTurbofanArchitecting software~\cite{bussemaker2024hidden}.
Through empirical evaluations and case studies, we demonstrate the effectiveness of our integrated approach in optimizing jet engine architecture design under real-world constraints, including hidden constraints.
This work contributes to advancing the field of system architecture optimization and offers valuable insights for future research in surrogate-based optimization techniques.

\end{abstract}

\section{Introduction}
\renewcommand*\footnoterule{}

\begin{figure}[H]
\begin{center}
\includegraphics[scale=0.45]{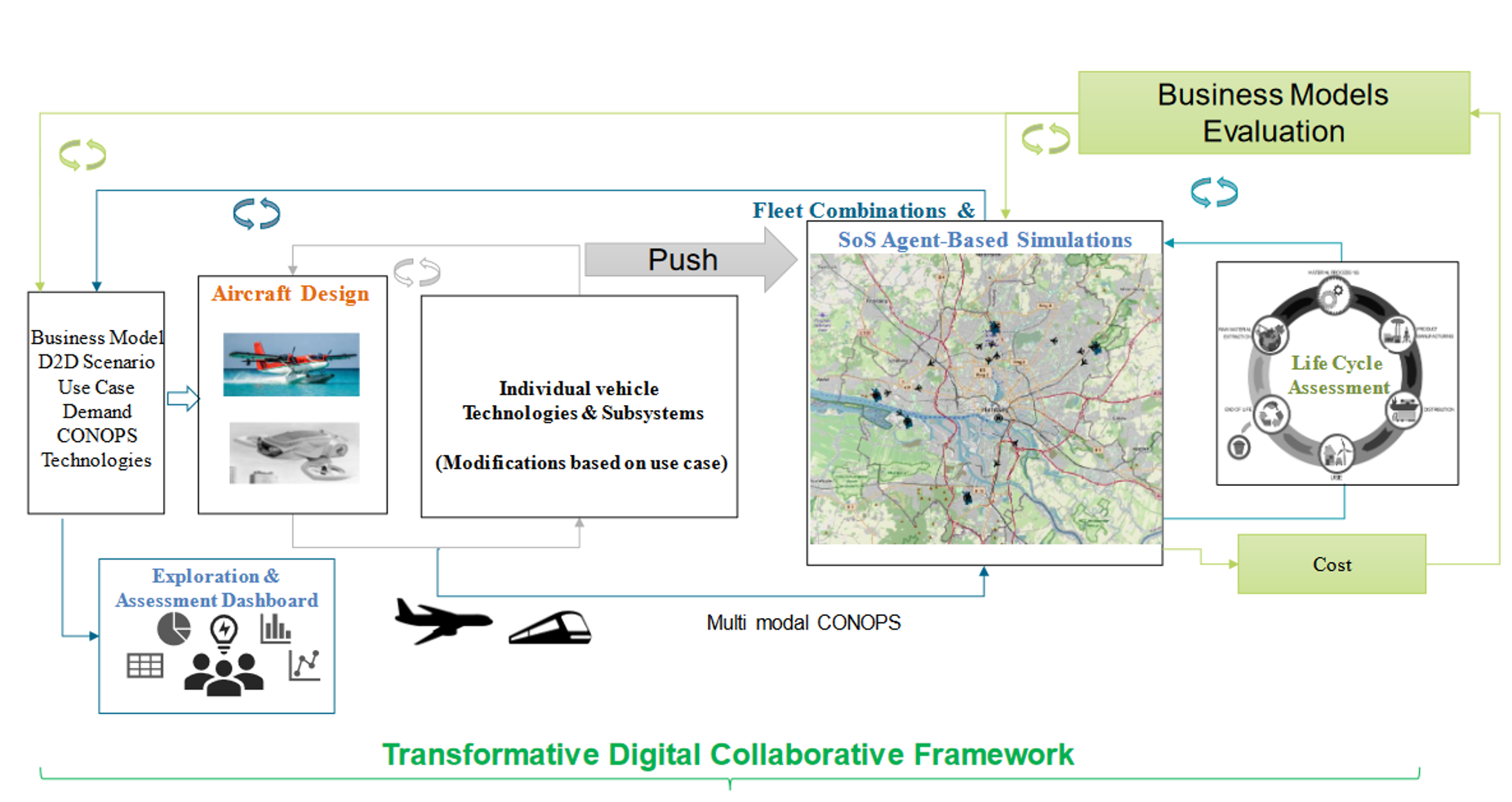}
\end{center}
\caption{COLOSSUS EU project Sustainable Intermodal Mobility use case.}
\label{fig:Adam}
\end{figure}

\label{sec:intro}
\lettrine{I}n the context of the European Union (EU) funded COLOSSUS project\footnote{\url{https://colossus-sos-project.eu/project/}} (Collaborative System-of-systems exploration of aviation products, services and business models) led by DLR, and in which both ONERA and DLR interact, several methods to analyze and optimize systems-of-system (SoS) are investigated. In particular, the first use case is to create a business model for sustainable 4D-intermodal mobility evaluating the concept for performance, competitiveness, environmental impact and life cycle footprint as shown in Figure~\ref{fig:Adam}. To do so, new aircraft configurations with a lower footprint on the environment (also known as Eco-aircraft design) have seen a resurgence of interest \cite{Eco-material,SEGO-UTB-Bombardier,AGILE} and, for really short-range passenger air transport, two types of aircraft are investigated besides other vehicles such as trains. The first aircraft is an all-electric vertical take-off and landing aircraft concept designed to improve 4D Intermodal Mobility. Multi-role considerations for this vehicle include its use in the Aerial Wildfire Fighting use case. The entry into service is assumed to be 2030. The second aircraft is a multi-Role Seaplane that is the first of its kind in CS23 class to be designed considering mobility and wildfire fighting requirements. Role-specific configurations of the seaplane are to be designed considering commonality, manufacturing, and operational requirements. The development of the multi-role seaplane is being explored within two distinct temporal horizons: the short-term scenario with an Entry into Service by 2035, and the long-term targeting EIS by 2050. %
\vspace{-0.5cm}

 The reminder of this paper is as follows.
Section~\ref{smt} introduces surrogate modeling for computationally expensive black-box functions. Using those surrogates, in  
Section~\ref{segomoe}, we describe the multi-objective constrained Bayesian optimization framework considering mixed hierarchical decision variables. Then, Section~\ref{adsg} presents the way architecture design problems are defined within the DLR software. Section~\ref{sbarchopt} focuses on optimization techniques within ONERA software to display joint software capabilities applied to an engine optimization.  Conclusions and perspectives are finally drawn in Section~\ref{sec:conclusion} including intermodal
mobility and short range aircraft optimization.

\section{Efficient surrogate models of computationally expensive black-box simulations}
\label{smt}

\begin{figure}[H]
\hspace{-1cm}
\includegraphics[clip=true,  height=5.0cm]{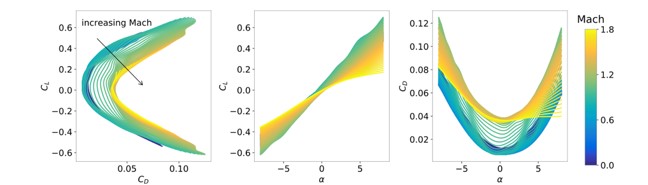}
\caption{Drag polar and aerodynamic properties for an  efficient supersonic air vehicle obtained from a surrogate model for different Mach speed and sweep angles~\cite[Figure 3]{Abbaqus}.}
\label{fig:Abbaqus}
\end{figure}

In this context, one targets to minimize the footprint on the environment of the aircraft using a Multidisciplinary Design Analysis (MDA)~\cite{Lambe2011,Lambe2012,Lambe2013}. This is an example of a computationally  expensive-to-evaluate (where  derivatives are not available) problem that could be encountered on industry. Therefore, it could be useful to use a surrogate model that reduces computational cost of while giving a good approximation of the black-box simulation.  The surrogate is often built from a small expensive-to-evaluat data, Design of Experiments (DoE), set of known configurations. 
An example of surrogate models in the context of aircraft design is given in~\figref{fig:Abbaqus}. 

Nevertheless, the process generally involves mixed continuous-categorical design variables. For instance, the size of aircraft structural parts can be described using continuous variables; in case of thin-sheet stiffened sizing, they represent panel thicknesses and stiffening cross-sectional areas. The set of discrete variables can encompass design variables such as the number of panels, the list of cross sectional areas or the material choices. \\
Moreover, variables hierarchy often occurs in variable-size problems or within problems featuring technological choices. For example, if a variable corresponds to the number of engines, the number of motors will vary accordingly. Similarly, if the chosen motor is electric, a battery needs to be selected whereas no battery is needed for a thermal motor. In consequence, some variables influence some others, leading to a hierarchy of variables.
In this work, we target to construct an inexpensive surrogate model $ \hat{f}$ for a black-box simulation $f$. 
The function $f$ is typically expensive-to-evaluate simulations with no exploitable derivative information.
Gaussian Processes (GP)~\cite{Jones98,Mockus,Rasmussen,forrester,Sasena,saves2021constrained}, also called Kriging models, are known to be a good modelling strategy to define response surface models. Namely,
we will consider that our unknown black-box  function $f$ is a realization of an underlying GP of mean $\mu^{f}$  and of standard deviation $s^f$, i.e.,
\begin{equation}
f \sim \hat{f}=\mathcal{G} \mathcal{P}\left(\mu^{f}, [s^f]^2\right). \label{eq:GP:f}\end{equation}

For a general problem involving categorical or integer variables, we proposed a dedicated GP in~\cite{Mixed_Paul} extended to high dimension in~\cite{Mixed_Paul_PLS} and to hierarchical variables in~\cite{saves2023smt}. Such models are available in open-source in 
SMT~\cite{SMT2019,saves2023smt}. SMT is developed in collaboration between ONERA, ISAE-SUPAERO, DLR, University of Michigan, University of California San Diego, Polytechnique Montréal and NASA Glenn. 

\section{Multi-objective constrained Bayesian optimization with mixed hierarchical decisions}
\label{segomoe}
\begin{figure}[H]
\centering
\includegraphics[width=0.8\textwidth]{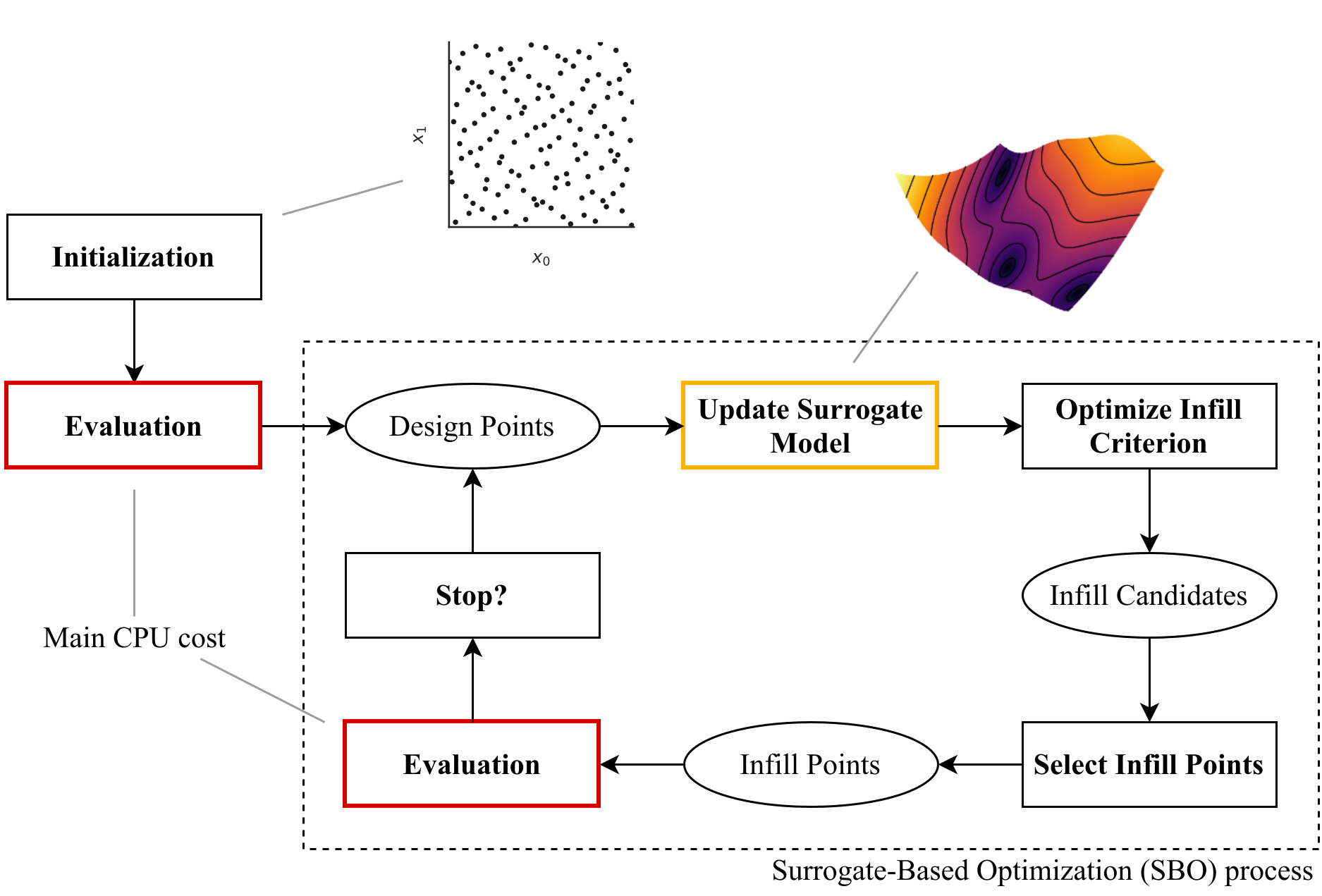}
\caption{The Bayesian optimization framework~\cite[Figure 2]{bussemaker2024system}.}\label{fig:sbo}
\end{figure}

Once the GP model is defined, we can use it for infill search and for optimizing an expensive black-box function through adaptive Bayesian Optimization (BO). In other words, we want to optimize a function by evaluating with as few costly evaluations as possible.
BO rely on GP models of the objective and constraint functions in order to efficiently determine the most interesting design point to evaluate next. Once the infill point have been evaluated using the expensive evaluation functions, the surrogate model is reconstructed and the process starts over, until some termination criterion has been reached. Figure~\ref{fig:sbo} visualizes the working principle of BO with, highlighted in red the costly evaluations whose number is limited to a certain computational budget and in orange the GP surrogate models from SMT for objective and constraint functions. 

For multi-objective optimization, which often occurs in air mobility or aircraft design cases, we extended BO in~\cite{grapin_constrained_2022}. The algorithm is based on 
SEGOMOE~\cite{bartoli:hal-02149236} which is developed by ONERA and ISAE-SUPAERO.  SEGOMOE builds a specific GP for each constraints and optimizes iterative an acquisition function subject to the surrogates of constraints. The approach used is based on the SEGO-UTB criterion developed in~\cite{SEGO-UTB} and that has been successfully applied to the BOMBARDIER BRAC aircraft in~\cite{SEGO-UTB-Bombardier}. The computer simulation can crash from an unpredictable error within the code and such crash problems are handled automatically with a GP classifier leading to a criterion named WB2S\_FE~\cite{tfaily2024bayesian,tfaily2023efficient}. 

\section{Architecture design problems definition}
\label{adsg}
\begin{figure}[H]
\centering
\includegraphics[scale=0.6]{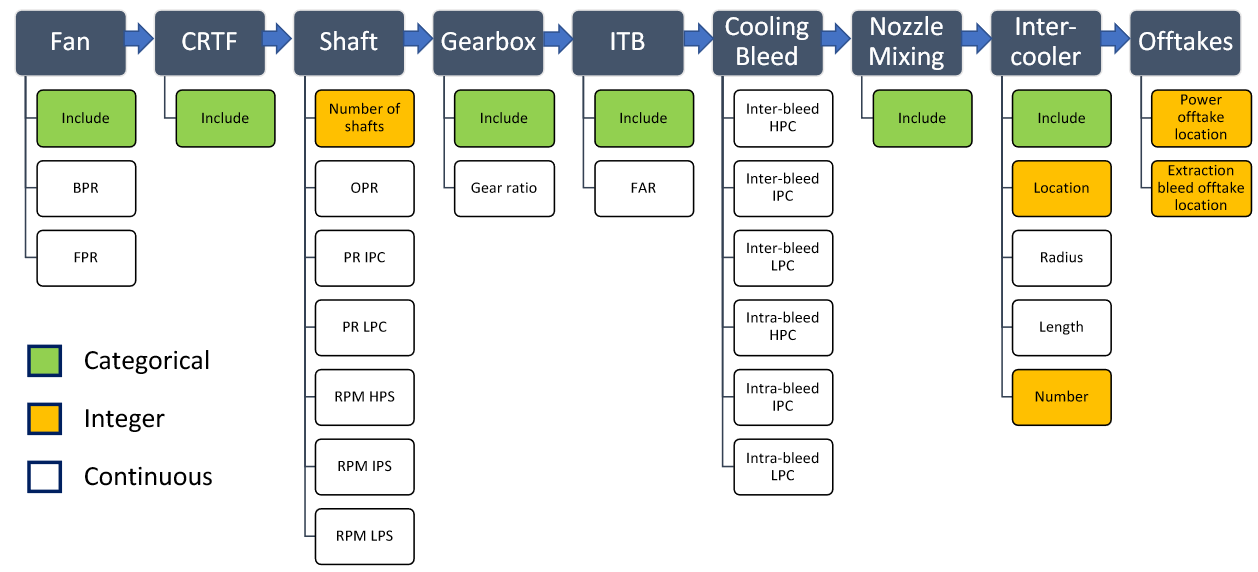}
\caption{Turbofan Architecting choices~\cite[Figure 7]{bussemaker2021system}.}
\label{fig:jta}
\end{figure}

At DLR, many works have been done to facilitate the definition of architecture design problems, in particular in Bussemaker \textit{et al.}~\cite{Architecture, bussemaker2022adore}. In particular, the Architecture Design Space Graph (ADSG) software allows to visualize architecture problems and the DLR Architecture Design and Optimization Reasoning Environment (ADORE) software allows its definition automatically.
Moreover, the open-source SBArchOpt software~\cite{bussemaker2023sbarchopt} interfaces optimizers such as SEGOMOE or SMARTy~\cite{bekemeyer2024recent} with architecture problems and is the keystone of this paper. Most problems interfaced are analytical but one developed in Bussemaker \textit{et al.}~\cite{bussemaker2021system} concerns an open-source architecture benchmark of jet engine in the open-source software OpenTurbofanArchitecting featuring hidden constraints~\cite{bussemaker2024hidden} with a failure rate of around 65\%. To validate our methods and software, this test case will be used. 
With the different design choices implemented in the benchmark problem, a total of 85 distinct engine architectures and approximately 1.3 million engine design points can be generated taking all discrete variables into account as illustrated in Figure~\ref{fig:jta}.  Verification and validation showed that the discipline results of the benchmark problem were very similar to actual engine data of the CFM LEAP-1C and the Pratt \& Whitney F100 engines.

\begin{figure}[H]
\centering
\includegraphics[scale=0.4]{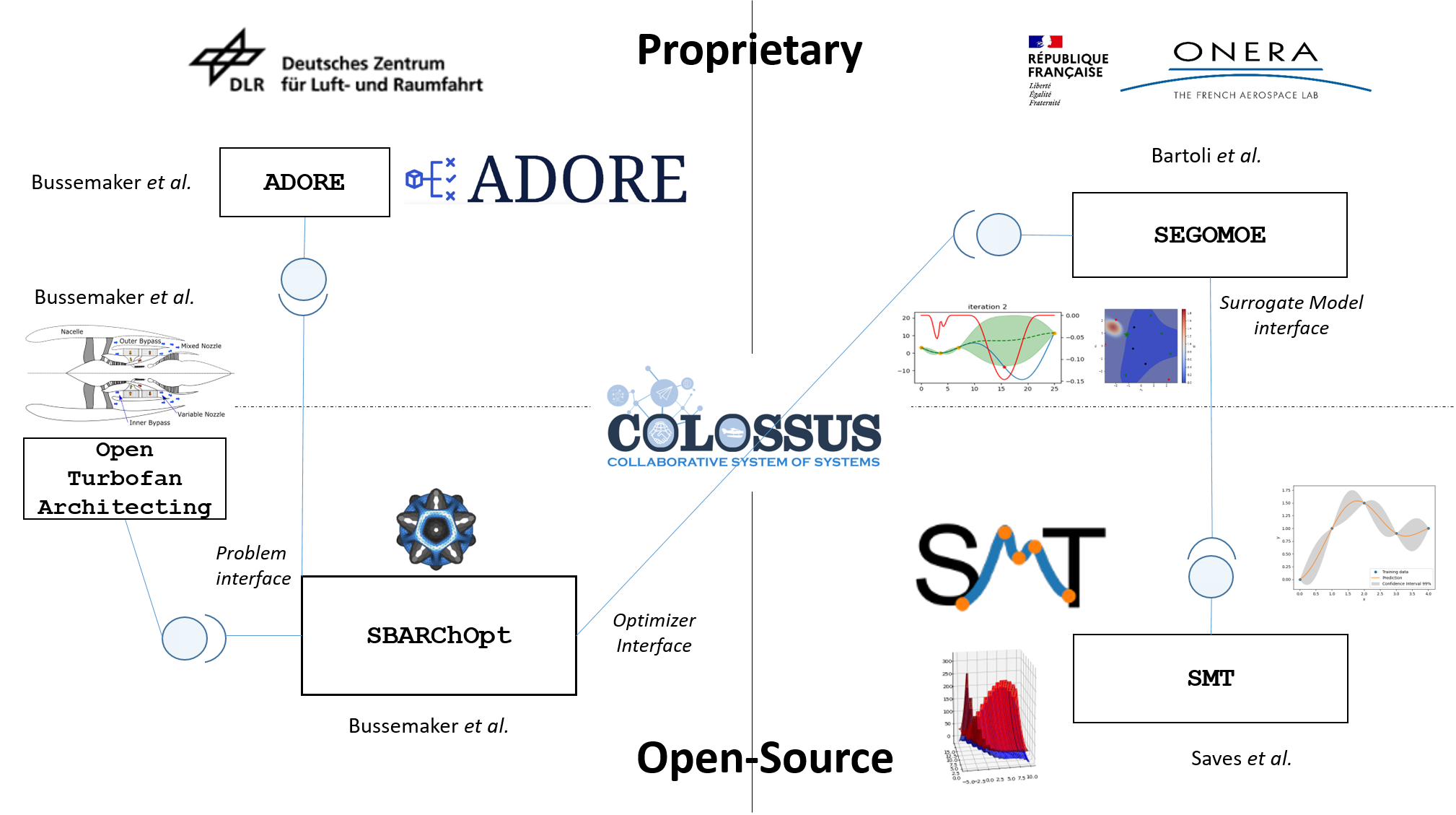}
\caption{The framework used in this paper with DLR and ONERA software.}
\label{fig:framework}
\end{figure}

From a software point of view, the overall final framework is described in Figure~\ref{fig:framework}.  DLR software is on the left and ONERA software on the right. Similarly, open-source software is at the bottom and proprietary software at the top. Note that SEGOMOE is not open-source but can be accessed through an ask-and-tell interface freely from the open-source software ONERA WhatsOpt~\cite{lafage2019whatsopt}. ADORE has been accessed through an interface and is available to all European partners~\cite{bussemaker2021system}.

\section{Architecture design problems optimization}
\label{sbarchopt}
In~\cite{bussemaker2021system}, a simple architecture problem with only 15 design variables and only conventional engine architectures leads to a reduced number of 15 distinct engine architectures and 216 engine design points. This optimization problem is mixed hierarchical and can be tackled by SEGOMOE thanks to the more recent advances in SMT 2.0~\cite{saves2023smt}.  The  optimization problem of Table~\ref{tab:optim_prob} is considered as in~\cite{bussemaker2024system}.

\begin{table}[H]
    \centering
   \caption{Definition of the turbojet engine optimization problem.}
   \small
   \resizebox{0.82\columnwidth}{!}{%
      \begin{tabular}{lllrr}
 & Function/variable & Nature & Quantity & Range\\
\hline
\hline
Minimize & Thrust-specific fuel consumption & cont & 1 &\\
 & \multicolumn{2}{l}{\bf Total objectives} & {\bf 1} & \\
\hline
with respect to & Bypass ratio & cont  &  1 &  $[2.0, 12.5] $ \\
& Fan pressure ratio & cont  &  1 & 
$[1.1, 1.8] $ \\
& Overall pressure ratio & cont  &  1 & $[1.1, 60.0] $ \\
& pressure ratio factor in shaft(s)* & cont  &  3 & $[0.1, 0.9] $ \\
&Rotation speed of shaft(s)* & cont  &  3 & $[1000, 20000] $ RPM \\

 & \multicolumn{3}{l}{* only active if the corresponding shaft is included} \\
 & \multicolumn{2}{r}{Total continuous variables} & {9} & \\ \cline{2-5}
& $n_\text{shafts}$ & discrete  &  1 & 
$ \{1,2,3\} $ \\
& Power offtake  & discrete  &  1 & 
$ \{1,2,3\} $ \\
& Bleed offtake & discrete  &  1 & 
$ \{1,2,3\} $ \\
 & \multicolumn{2}{r}{Total discrete variables} & {3} & \\ \cline{2-5}
& Include fan? & cat  &  1 &  $\{True, False\} $ \\
& Include gearbox?$^\dag$ & cat  &  1 &  $\{True, False\} $ \\
& Mixed nozzle?$^\dag$ & cat  &  1 &  $\{True, False\} $ \\
 & \multicolumn{3}{l}{$^\dag$only active if the fan is included} \\

& \multicolumn{2}{r}{ Total categorical variables} & { 3} & \\ \cline{2-5}
 & \multicolumn{2}{r}{\bf Total relaxed variables} & {\bf 18} & \\
 \hline
subject to &  \ $M_{\text{jet}}$ \textless \  1 & cont & 1 \\
& Sum of pressure ratio factor(s) \textless \  0.9   & cont &1\\
& maxima of pressure ratio factor(s) in shaft(s) \textless \  15   & cont &3\\
 & {\textbf{Total  constraints}} &  & {\textbf{5}} & \\
\hline
\end{tabular}
}
   \label{tab:optim_prob}
\end{table}

The problem introduced above includes several levels of activation hierarchy: bypass ratio $BPR$, fan pressure ratio $FPR$, gearbox inclusion $IncludeGearbox$ and mixed nozzle selection $MixedNozzle$ are only active if the fan is included ($IncludeFan = True$); the gear ratio $GearRatio$ is only active if $IncludeGearbox = True$; and shaft-related pressure ratio factors $PR_{factor,i}$ and rotational speeds $RPM_i$ are only active if a sufficient number of shafts is selected.
The power offtake $PowerOfftake$ and bleed offtake $BleedOfftake$ selections are always active, however are value-constrained by the available shafts.
The problem additionally features 5 design constraints, constraining the output jet Mach number $M_{jet}$ and pressure ratio distributions over the selected compressor stages ($PR_{factor,sum}$ and $PR_{max,i}$). 

\begin{figure}[H]
\centering
\includegraphics[width=0.92\textwidth]{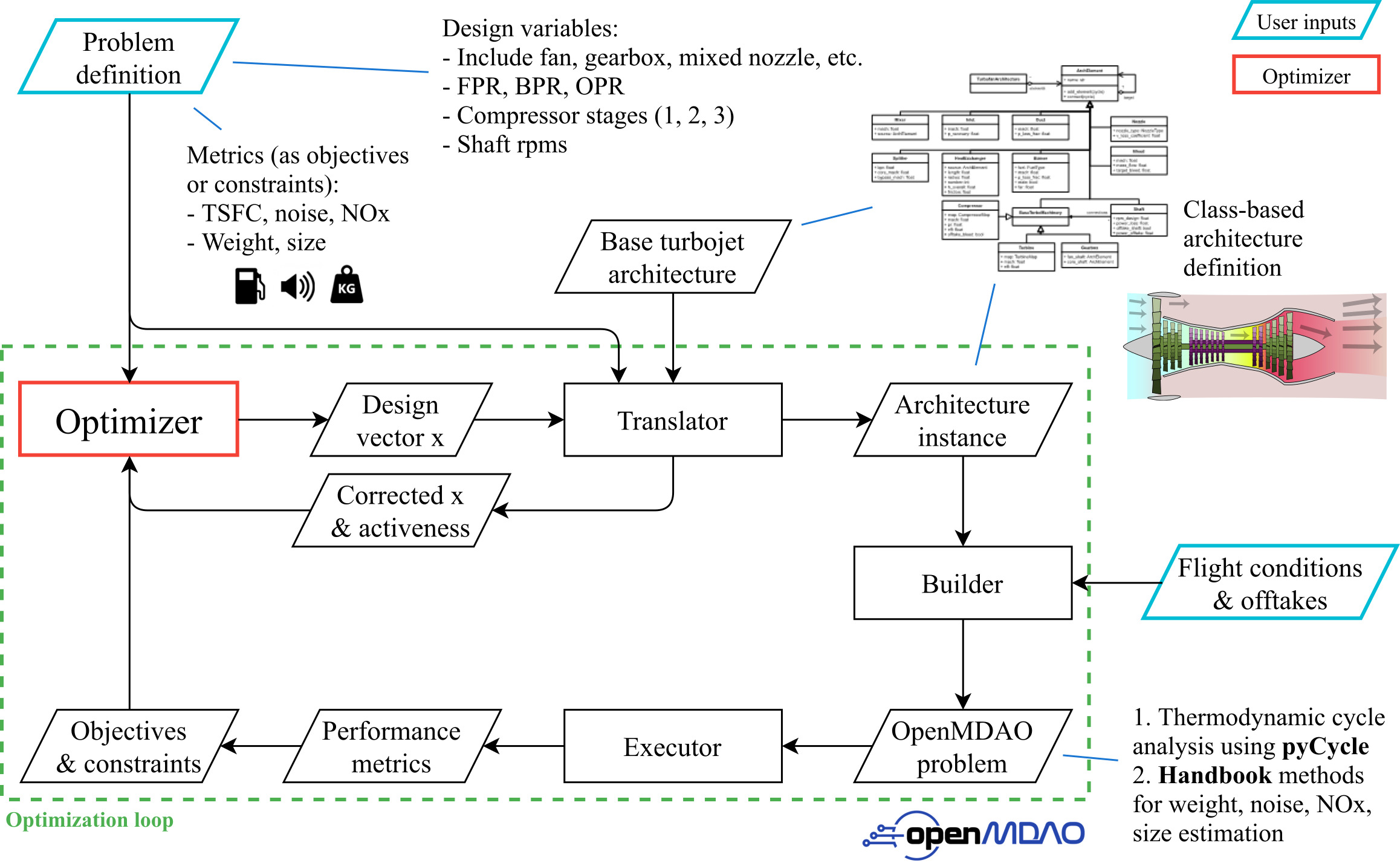}

\caption{Turbofan architecting code overview~\cite[Figure 5]{bussemaker2024system}.}
\label{fig:jet_eng_prob}
\end{figure}

Additionally, the underlying thermodynamic cycle analysis and sizing code does not always converge, leading to a hidden constraint being violated in approximately 50\% of design points generated in a random DoE.  Handbook methods are added to calculate additional metrics such as noise level, NOx emissions, weight, and size. Note that thermodynamic cycle analysis takes between 1 and 5 minutes to complete and is representative of an expensive-to-evaluate black-box engineering problem.
The corresponding code overview is given in Figure~\ref{fig:jet_eng_prob} and more information are given in~\cite{bussemaker2024system}. 

\begin{figure}[H]
\centering
 \subfloat[Jet engines compared for 4 technological choices.
 ]{   
      \centering 
	\includegraphics[clip=true,  width=17cm]{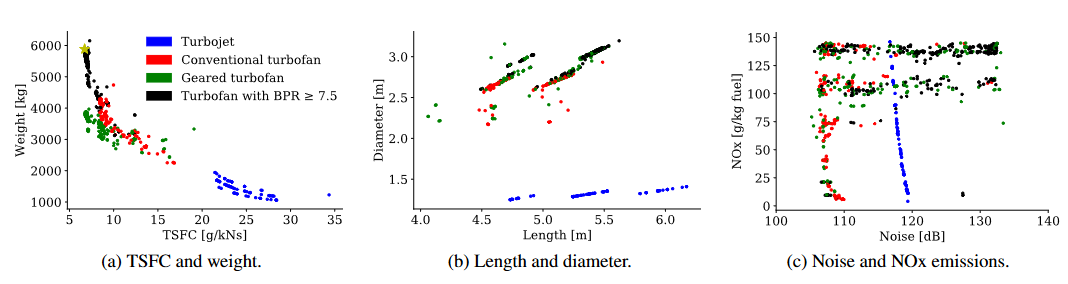}} \\
  \subfloat[Jet engines compared with 1 to  3 shafts.]{
      \centering 
	\includegraphics[clip=true,  width=17cm]{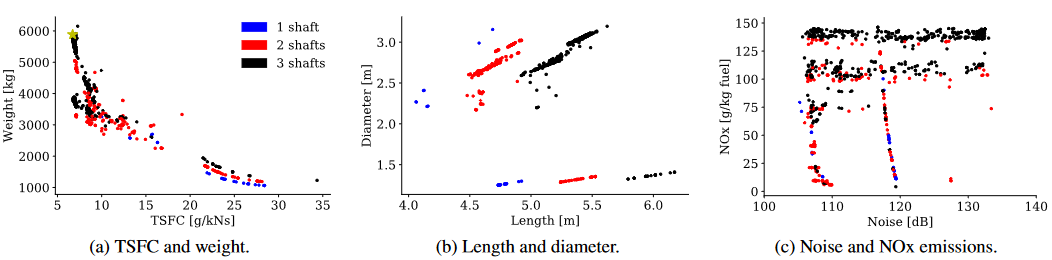}}
\caption{Jet engine architectures analyses.}
\label{fig:analysis}
\end{figure}

For every of the 216 different engines, up to 9 continuous variables are to optimize.  All these architectures have been analyzed (Thrust Specific Fuel Consumption/Weight, Length/Diameter, Nox emissions/Noise) in Figure~\ref{fig:analysis}. In particular, the minimal TSFC is of $6.633$ g/kNs. 
The comparisons show that turbojets have higher TSFC and lower weight and diameter than turbofans. Increasing shaft numbers reduces TSFC but increases length and weight due to additional compressors. Geared turbofans achieve low TSFC by optimizing compressor and fan speeds while maintaining low weight. UHBR turbofans reduce TSFC but increase weight due to larger engine diameter.

We use the problem implemented in SBArchOpt as SimpleTurbofanArch. The effectiveness of the Evolutionary algorithm Non-Sorting Genetic Algorithm 2~\cite{nsga2} (NSGA-II) serve as a baseline for comparison in Table~\ref{tab:jet_engine_opt}.

\begin{table}[H]
\caption{Comparison of optimal TSFC values achieved for the jet engine problem. The number of function evaluations $N_{fe}$ (i.e., the computational budget) used to get the optimal value is also mentioned.
}
\centering
\label{tab:jet_engine_opt}
\begin{tabular}{llc}
\toprule
Algorithm  & $N_{fe}$ & TSFC [g/kNs]  \\
\midrule
SEGOMOE  & 60 &  6.679\\
SEGOMOE  & 300 &  6.673\\
\hdashline
NSGA-II  & 300  & 7.455 \\
NSGA-II  & 1000  & 6.7  \\
\bottomrule
\end{tabular}
\end{table}

Table~\ref{tab:jet_engine_opt} shows highly improved optimization results with SEGOMOE algorithm compared to NSGA-II, in particular with a small computational budget of 60 points, outperforming the evolutionary algorithm that need 17 times more evaluations for the same results.  The strength of Bayesian optimization is clearly displayed and has been extended to many multi-objective aircraft design problems in~\cite{saves24} and in particular, it served in previous works to reduce the fuel consumption of hybrid-electric aircraft~\cite{SciTech_cat}.

\section{Conclusion}
\label{sec:conclusion}
In conclusion, this paper has effectively demonstrated the synergistic capabilities of DLR and ONERA software platforms, showcasing their integration abilities for efficient modeling and optimization of complex black-box aircraft problems using  limited computational resources. 
In particular, we leverage the SEGOMOE and SMT ONERA software to efficiently model and optimize expensive-to-evaluate aircraft architecture black-box problems. Those problems were defined thanks to the DLR ADORE OpenTurbofanArchitecting tools. Notably, we showcase the ability to collaborate in an intermodal fashion, we relied on DLR open-source software SBArchOpt to underscore our joint commitment for innovation and collaboration. This in particular offers a comprehensive solution to intricate design problems with respect to both modeling and optimization. Future works will include even more collaboration in an online DLR platform named RCE~\cite{boden2021rce} for both wildfire fighting and intermodal on-demand mobility concept of operations integration and aircraft design optimization as presented in Section~\ref{sec:intro}.




\section*{Acknowledgements}
This work is part of the activities of ONERA - ISAE - ENAC joint research group. The research presented in this paper has been performed in the framework of the COLOSSUS project (Collaborative System of Systems Exploration of Aviation Products, Services and Business Models) and has received funding from the European Union Horizon Programme under grant agreement n${^\circ}$101097120.
\bibliography{main}

\end{document}